\documentclass[hidelinks]{isprs}
\usepackage{subfigure}
\usepackage{setspace}
\usepackage{geometry} 
\usepackage{epstopdf}
\usepackage[labelsep=period]{caption}  

\usepackage{hyperref}
\usepackage{graphicx}
\usepackage[export]{adjustbox}
\usepackage{amsmath,amsfonts,amssymb}
\usepackage[acronyms]{glossaries}
\usepackage[sort]{natbib} 
	\renewcommand{\refname}{REFERENCES} 
\usepackage{color, soul}
\usepackage{diagbox}	
\usepackage{enumitem}
\usepackage{gensymb}
\usepackage[nameinlink, capitalize]{cleveref}
\usepackage{hhline}

\usepackage[percent]{overpic}

\DeclareMathOperator*{\argmin}{arg\,min}
\DeclareMathOperator*{\absmax}{abs\,max}
\usepackage{xspace}

\newcommand{\ie}{i.e.\ }

\newcommand{\eg}{e.g.\ }

\newcommand{\cf}{cf.\ }

\newcommand{\wrt}{w.r.t.\ }

\newcommand{\approxi}{approx.\ }

\newcommand{\m}{\,m\xspace}

\newcommand{\kilo}{\,k\xspace}

\newcommand{\kg}{\,kg\xspace}

\newcommand{\ms}{\,ms\xspace}

\newcommand{\Hz}{\,Hz\xspace}

\newcommand{\MHz}{\,MHz\xspace}

\newcommand{\fps}{\,fps\xspace}

\newcommand{\px}{\,px\xspace}

\newcommand{\TULIPP}{\textsc{Tulipp}\xspace}

\newcommand{\KITTI}{KITTI\xspace}

\begin{document}

\title{Real-time on-board obstacle avoidance for UAVs \mbox{based on embedded stereo vision}}

\author{
B. Ruf\textsuperscript{a,b}, S. Monka\textsuperscript{a}, M. Kollmann\textsuperscript{a}, M. Grinberg\textsuperscript{a}
}

\address{
\textsuperscript{a}Fraunhofer IOSB, Video Exploitation Systems, Karlsruhe, Germany -\\ \{boitumelo.ruf, sebastian.monka, matthias.kollmann, michael.grinberg\}@iosb.fraunhofer.de\\
\textsuperscript{b}Institute of Photogrammetry and Remote Sensing, Karlsruhe Institute of Technology, \\Karlsruhe, Germany - \{boitumelo.ruf\}@kit.edu
}


\commission{I, }{I} 
\workinggroup{} 
\icwg{I/II}   


\newacronym{UAVs}{UAVs}{unmanned aerial vehicles}
\newacronym{UAV}{UAV}{unmanned aerial vehicle}
\newacronym{WTA}{WTA}{winner-takes-it-all}
\newacronym{SGM}{SGM}{semi-global matching}
\newacronym{GPGPU}{GPGPU}{general purpose computation on a GPU}
\newacronym{COTS}{COTS}{commercial off-the-shelf}
\newacronym{TULIPP}{TULIPP}{Towards Ubiquitous Low-power Image
Processing Platforms} 
\newacronym{MAVLink}{MAVLink}{Micro Air Vehicle Link}
\newacronym{SAD}{SAD}{sum of absolute differences}
\newacronym{CT}{CT}{Census Transform}
\newacronym{RoI}{RoI}{region of interest}
\newacronym{LiDAR}{LiDAR}{light detection and ranging}
\newacronym{HLS}{HLS}{high-level synthesis}

\abstract{
In order to improve usability and safety, modern \gls*{UAVs} are equipped with sensors to monitor the environment, such as laser-scanners and cameras. One important aspect in this monitoring process is to detect obstacles in the flight path in order to avoid collisions. 
Since a large number of consumer \gls*{UAVs} suffer from tight weight and power constraints, our work focuses on obstacle avoidance based on a lightweight stereo camera setup. 
We use disparity maps, which are computed from the camera images, to locate obstacles and to automatically steer the UAV around them.
For disparity map computation we optimize the well-known \gls*{SGM} approach for the deployment on an embedded FPGA.
The disparity maps are then converted into simpler representations, the so called U-/V-Maps, which are used for obstacle detection.
Obstacle avoidance is based on a reactive approach which finds the shortest path around the obstacles as soon as they have a critical distance to the UAV.
One of the fundamental goals of our work was the reduction of development costs by closing the gap between application development and hardware optimization.
Hence, we aimed at using \gls*{HLS} for porting our algorithms, which are written in C/C++, to the embedded FPGA.
We evaluated our implementation of the disparity estimation on the \KITTI Stereo 2015 benchmark.
The integrity of the overall real-time reactive obstacle avoidance algorithm has been evaluated by using Hardware-in-the-Loop testing in conjunction with two flight simulators.
}

\maketitle

\glsresetall 
\glsunset{UAV} 

\section{INTRODUCTION}
\label{sec:intro}
\sloppy

In recent years, the use of \gls*{UAVs} in different markets has increased significantly. Currently, most important markets are aerial video/photography, precision farming and surveillance/monitoring. Ongoing technical development, \eg in size, endurance and usability, enables the use of \gls*{UAVs} in even more areas. First prototypes for delivering goods or even for transporting people are already available.

In order to improve usability and safety, \gls*{UAVs} are equipped with sensors to monitor the environment, such as laser-scanners and cameras. One important aspect in this monitoring process is to detect obstacles in the flight path in order to avoid collisions. With the increasing performance of state-of-the-art algorithms in combination with modern hardware, camera systems are typically more practical than laser-scanners in performing this task, especially in terms of costs, weight and power-consumption, inherent to most \gls*{COTS} \gls*{UAVs}.

The \TULIPP\footnote{\url{http://tulipp.eu/}} project \citep{Kalb2016tulipp}, funded by the European Commission under the H2020 Framework Programme for Research and Innovation, aims to remedy these restrictions by setting up an ecosystem for a low-power image processing platform, consisting of a exemplary hardware instantiation, an energy-aware tool chain \citep{Sadek2018sthem} and a high performance real-time operating system \citep{Paolillo2015hipperos}. The Xilinx Zynq Ultrascale+ was selected  as a main processing unit for the hardware instantiation since its combination of an embedded quad-core CPU and a FPGA provides high performance at low power.

In order to validate the developed components, the project defines three use cases in three different domains. One of these use cases shall enhance the development of modern \gls*{UAVs} by providing a system for a real-time obstacle avoidance based on a lightweight and low-cost stereo camera setup. 
Hereby, the cameras are orientated in the direction of flight.
We use disparity maps, which are computed from the camera images, to locate obstacles and to automatically steer the \gls*{UAV} around them.
As in the other use cases of the \TULIPP project, the focus of our work lies, among other things, on a user-friendly development and optimization of image processing algorithms for embedded hardware. The aim is to close the gap between algorithmic development and hardware optimization and to facilitate the reduction of time-to-market, development and rework costs.

\begin{figure*}[!ht]%
	\centering
	\includegraphics[width=\textwidth]{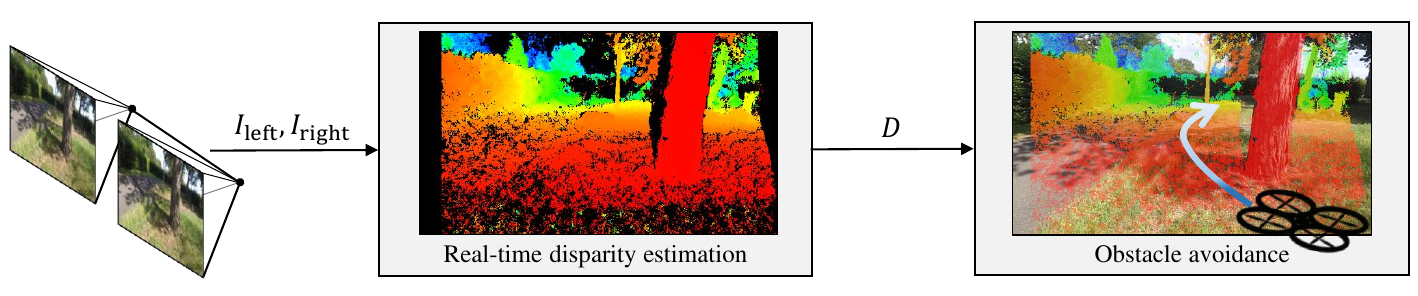}
	\caption{Overview of our system for real-time on-board obstacle avoidance based on embedded stereo. We use a stereo camera system to acquire two input images $I_\mathrm{left}$ and $I_\mathrm{right}$. These are passed to the first processing stage, where a disparity map $D$ is computed in real-time on an embedded FPGA. This disparity map is used by the subsequent step to compute and execute an evasion manoeuvre in order to avoid obstacles in the flight path.}
	\label{fig:overview_sys}
\end{figure*}

The main contribution presented in this paper is a system for on-board obstacle avoidance that
\begin{itemize}
\item employs the \gls*{SGM} algorithm \citep{Hirschmueller2008} that is synthesized from C/C++ code and is deployed on an embedded FPGA in order to compute disparity information from a stereo camera setup,
\item performs a reactive obstacle avoidance based on the precomputed depth information by calculating the shortest path around the obstacle and executing the evasion manoeuvre on the \gls*{UAV}, and
\item runs in real-time on an embedded Xilinx Zynq Ultrascale+, which provides a good performance-to-watt ratio and is therefore a suitable choice when it comes to low-power real-time processing.
\end{itemize}

The paper is structured as follows: in \Cref{sec:related_work} we give a brief overview of previous work done in the field of autonomous obstacle avoidance for \gls*{UAVs} and the advancements achieved in embedded stereo vision. 
This is followed by a detailed description of our approach in \Cref{sec:methodology}, which involves an algorithm of real-time embedded disparity estimation from stereo imagery and a system for obstacle avoidance based on the computed disparity information. In \Cref{sec:eval} we present the experimental results that were achieved. \Cref{sec:conclusion} concludes the paper by a short summary and an outlook.

\section{RELATED WORK}%
\label{sec:related_work}
\sloppy

In past years a substantial amount of work has been done in the field of navigating mobile robots securely through a known or unknown environment. 
Ground operating robots, or slow moving and heavy \gls*{UAVs}, typically use \gls*{LiDAR} technology to conceive the surrounding environment \citep{Bachrach2012}. 
\cite{Scherer2005} used a heavy helicopter able to carry maximum payload of 29\kg and flying with speeds of up to 10\m/s. 
The \gls*{LiDAR} system has a range up to 150\m, which is suitable given the speeds and the mass of the system.
However, for fast reactions in dynamic or cluttered scenes, heavy \gls*{LiDAR} systems are not practical. 
They consume significantly more power than a camera-based system, thus requiring a corresponding power supply, making them heavier and sluggish. 

Due to higher speeds and more agile movements, \gls*{UAVs} need fast, light and energy-efficient solutions for robust obstacle avoidance. 
Systems relying on a single camera typically use optical flow information, computed between two consecutive frames to perceive the structure of the environment. 
The main advantage of such structure-from-motion approaches is that they are lightweight and energy-efficient and can therefore be used for small \gls*{UAVs}. 
\cite{Lee2010} proposed a probabilistic method of computing the optical flow for more robust obstacle detection. 
\cite{Ross2015} used a data driven optical flow approach learning the flight behaviour from a human expert. 
Nonetheless, since the information gained by a single camera system is limited, such systems are typically infeasible for robot navigation.

The use of active RGB-D systems, such as structured-light cameras or time-of-flight sensors, provides both colour images and per-pixel depth estimates. The richness of this data and the recent development of low-cost sensors make them very popular in mobile robotics research \citep{Bachrach2012}. However, such sensors suffer from a limited range or they require homogeneous light conditions, making them impractical for the use in outdoor environments. 

In recent years the performance of image-based depth estimation algorithms has increased significantly. These achievements in conjunction with modern hardware, make passive RGB-D vision sensors, such as stereo camera systems feasible for indoor and outdoor environments. Depending on the baseline of the cameras, \ie the distance at which they are mounted on the rig, and their image resolution, such sensor system can be used for close range as well as far range depth estimation \citep{Gallup2008variable}. Numerous studies have shown the strength of stereo vision \wrt on-board obstacle avoidance for \gls*{UAVs} \citep{Barry2015, Oleynikova2015, Schmid2013stereo}. 

When it comes to real-time stereo vision, the \gls*{SGM} approach \citep{Hirschmueller2008} has emerged as one of the most widely used algorithm, especially for embedded systems. This is not only because of its accuracy and performance, but also due to the fact that it is well suited for parallelization on many architectures. \cite{Spangenberg2014} as well as \cite{Gehrig2010} optimized the \gls*{SGM} algorithm to run on a conventional CPU, reaching 12 and 14 fps respectively at a VGA image resolution. The utilization of \gls*{GPGPU} can lead to faster processing speeds at higher image resolutions, as the work \citep{Banz2011} and \citep{Haller2010} show.

Nonetheless, since \gls*{GPGPU} is typically not very energy-efficient, most adaptations of the \gls*{SGM} algorithm for embedded systems were done for the FPGA architecture. 
To mention a few, \cite{Barry2015} developed an FPGA based stereo vision sensor running with 120\fps at an image resolution of 320$\times$240 pixels with the \gls*{SGM} algorithm.
\cite{Hofmann2016} deployed the SGM algorithm on an Xilinx Zynq 7000 FPGA, achieving 32\fps on an image of 1280$\times$720 pixels.
Even frame rates of up to 197\fps can be achieved at a similar frame size depending on the performance of the FPGA, as \citep{Li2017} shows.

However, in order to achieve such performance, a thorough optimization of the algorithm is required, which typically involves the use of hardware specific APIs such as VHDL or Verilog.
In contrast, we aim at using \gls*{HLS} for porting our algorithm, which is written in C/C++, to the FPGA.
This is done specifically because one of the fundamental goals of the \TULIPP project is the reduction of development costs by closing the gap between application development and hardware optimization.
In doing so, we accept some potential performance loss.

\section{METHODOLOGY}%
\label{sec:methodology}
\sloppy

The real-time on-board obstacle avoidance system is based on disparity images of a calibrated stereo camera rig, which is mounted onto a \gls*{UAV} and is orientated in the direction of flight. 
These disparity images, which encode the structure of the environment in front of the \gls*{UAV}, are then used to detect objects in the flight path and to avoid a collision in further steps. 
The processing is done on-board in real-time and can be partitioned into two parts as depicted in \Cref{fig:overview_sys}, namely: real-time disparity estimation and obstacle avoidance. 
While our algorithm for disparity estimation is optimized for the execution on the FPGA, the detection of obstacles and the computation for the evasion manoeuvre is run on a general-purpose CPU. 
We use the \gls*{MAVLink} protocol to control the \gls*{UAV} and execute the evasion manoeuvre. 
In the following, both parts of our system will be discussed in more detail.

\begin{figure}[!ht]%
	\centering
	\includegraphics[width=0.6\columnwidth]{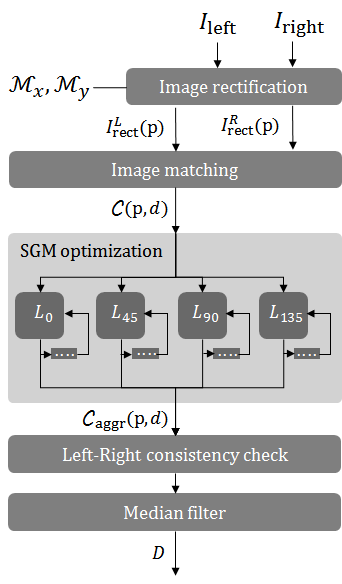}
	\caption{Schematic overview of our processing pipeline for real-time disparity estimation of a FPGA. Within the \gls*{SGM} optimization the costs are aggregated along 4 concentric paths, \ie $L_{0}$ - horizontal from left to right, $L_{45}$ - diagonal to bottom right, $L_{90}$ - vertical from top to bottom, $L_{135}$ - diagonal to bottom left. The aggregation costs are buffered in order to accumulate them while iterating of the image data.}
	\label{fig:overview_disp}
\end{figure}

\subsection{Real-time disparity estimation}
%
A stereo camera setup typically consists of two cameras which are mounted on a fixed rig with a baseline $b$ and orientated in a way that their optical axes are aligned in parallel.
Due to the slightly different viewing perspectives, a scene point $\mathrm{M}$ is projected onto different positions $\mathrm{m}_1$ and $\mathrm{m}_2$ in the images of both cameras.
When the camera parameters are known, finding corresponding pixels in the images of the cameras allows to compute the depth of the corresponding points and to reconstruct their 3d positions.
For correct scene reconstruction this has to be done for all pixels of the camera images.

In general, a correspondence search for a pixel in the image of one camera in the image of the second camera has to be done along the so called epipolar curve, which corresponds to the viewing ray of the first camera in the image of the second camera (epipolar constraint).
If it is possible to compensate for the camera distortions, the curve becomes a straight line.
Another simplification can be achieved through rectification of the images, i.e. transformation of both images onto a common image plane.
As a result, the epipolar lines coincide and corresponding image points of both rectified images turn out to lie in the same image row.
The displacement of the image points $\mathrm{m}_1$ and $\mathrm{m}_2$ of the scene point $\mathrm{M}$, which is called disparity, is then defined as
$d=\left| {\mathrm{m}_x}_1 - {\mathrm{m}_x}_2 \right|$.
Together with the known calibration parameters of the stereo rig, the disparity allows to reconstruct depth of the corresponding scene point.

Finding pixel-wise matches between two input images and thus the corresponding disparities leads to a disparity image.
Hereby, the correctness of the result greatly depends on the matching function and the optimization strategies that are used. 
The optimization strategies are typically categorized into local and global approaches.
While local methods rely on a fixed neighbourhood around a central pixel, global methods optimize an energy function on the full image domain.
Due to the finite window size, local methods are very efficient but often lack in accuracy \citep{Scharstein2002}.

In the past decade the proposed \gls*{SGM} approach \citep{Hirschmueller2008} demonstrated a good trade-off between runtime and accuracy in numerous studies. 
It uses dynamic programming to optimize a global energy function by aggregation the cost of each pixel $\mathrm{p}$ given the disparity $d_\mathrm{p}$ along multiple one-dimensional paths that cover the complete image. 
Hereby, each path computation is an independent sub-problem, making \gls*{SGM} suitable for highly parallel execution on dedicated hardware. 
While it is initially proposed to use 16 concentric paths, most studies done on conventional hardware only use eight paths in order to avoid subpixel image access. 
However, due to the limited amount on memory on embedded FPGAs and their strengths in processing data streams, it is common practice to only use four paths when porting the algorithm onto such hardware (\cf \Cref{sec:sgm}). 
Studies show that a reduction from eight to four paths lead to a loss in accuracy of 1.7\% while greatly increasing the performance of the algorithm \citep{Banz2010}.

We have adopted and optimized this approach for real-time image-based disparity estimation on embedded hardware. Our full algorithmic chain is comprised of the following steps:

\begin{itemize} [topsep=0pt, partopsep=0pt, itemsep=0pt]
\item Image rectification
\item Image matching 
\item \gls*{SGM} optimization
\item Left-Right consistency check
\item Median filter
\end{itemize}

These steps are pipelined in order to efficiently process the pixel stream produced by the cameras (cf. \Cref{fig:overview_disp}). We use FIFO buffers to pass the data from one step to the next without using too much memory. Everything is implemented in C/C++ and then synthesized into VHDL code with \gls*{HLS} to run on the FPGA.

\subsubsection{Image rectification}

Each pair of input images, for which the disparity estimation is to be done, needs to be rectified. However, the calibration parameters can be precomputed so that each image pair of the sequence is transformed in the same manner. 
We use a standard calibration routine \citep{Zhang2000flexible} to compute two rectification maps for each image $\mathcal{M}_\mathrm{x}, \mathcal{M}_\mathrm{y} \in \mathbb{N}_0^{W \times H}$, which hold for each pixel $\mathrm{p} = (x,y)$ in the rectified image $I_\mathrm{rect}$ the coordinates of the corresponding pixel in the unrectified image $I$: $I_\mathrm{rect}\left(\mathrm{p}\right) =  I\left(\mathcal{M}_\mathrm{x}\left(\mathrm{p} \right), \mathcal{M}_\mathrm{y}\left(\mathrm{p} \right)\right)$. 
We have implemented a line buffer to store $n = \mathrm{\absmax}\left(\mathcal{M}_\mathrm{y}\right)$ lines of the input pixel stream in order to buffer enough pixel data to compute the rectified image.

\subsubsection{Image matching}
In the next stage of the pipeline the rectified images $I_\mathrm{rect}^\mathrm{L}, I_\mathrm{rect}^\mathrm{R}$ of the two cameras are used to perform a pixel-wise image matching by computing a similarity score $\Phi$ between each pixel in both images given the expected disparity range with a maximum disparity $d_\mathrm{max}$. This will generate a three dimensional cost volume $\mathcal{C} \in \mathbb{N}_0^{W \times H \times d_\mathrm{max}}$ in which each cell contains the pixel-wise matching cost for a given disparity $d$: 
\begin{equation}
\label{eq:img_match}
\begin{aligned}
	\forall d \in \mathbb{N}_0,& 0 \leq d < d_\mathrm{max}: \\&\mathcal{C}(\mathrm{p}, d) = \Phi(I_\mathrm{rect}^\mathrm{L}(\mathrm{p}), I_\mathrm{rect}^\mathrm{R}((\mathrm{p}_x - d, \mathrm{p}_y)).
\end{aligned}
\end{equation}

Due to the rectification done in the previous step, the search space for similar pixels in both images is reduced to the scan lines of the images. A relative translation of the second camera to the right, results in a displacement of the corresponding pixel to the left of the reference pixel.
Hence, the occurrence of a pixel correspondence in the right image is restricted to the left side of the original position of the reference pixel in the left image. Since the pixel data of both images are processed as streams, $m = d_\mathrm{max}$ pixels of the right image have to be buffered in order to allow a search for similar pixels in already processed data.

There exist a numerous number of appropriate cost functions to measure the similarity between two pixels, which work on different information a pixel can provide. 
We have employed a simple pixel-wise \gls*{SAD}, as well as the non-parametric Hamming distance of the \gls*{CT} \citep{Zabih1994}, due to its popularity in real-world scenarios. 
Again, due to the local support regions, these cost function require the use of line buffers in order to use pixel data in the local neighbourhood. 

\subsubsection{SGM optimization}
\label{sec:sgm}
The previously computed cost volume $\mathcal{C}$ holds the plausibilities of disparity hypotheses for each pixel $\mathrm{p}$ in the left image. This could already be used to extract a disparity image by finding the \gls*{WTA} solution for each $\mathrm{p}$, \ie the disparity with the minimum costs. However, due to ambiguities in the matching process this would not generate a suitable result. Therefore an optimizations strategy is typically applied to regularize the cost volume $\mathcal{C}$ and remove most of the ambiguities. As already stated, we use the \gls*{SGM} optimization scheme, due to its suitability in parallelization and hence, its suitability for use on embedded hardware.  

In its initial formulation the \gls*{SGM} approach optimizes an energy function by aggregating the cost for each pixel $\mathrm{p}$ in $\mathcal{C}$ given disparity $d_p$ along concentric paths that center in $\mathrm{p}$. 
As stated by \Cref{eq:sgm}, the costs of each neighbourhood pixel $\mathrm{q} \in \mathcal{N}_\mathrm{p}$ along the aggregation paths is penalized with a small penalty $P_1$ if its disparity $d_\mathrm{q}$ differs by one pixel from the disparity $d_\mathrm{p}$. 
If the disparity change between $d_\mathrm{p}$ and $d_\mathrm{q}$ is larger than $\pm$1, a greater penalty $P_2$ is added to the cost of pixel $\mathrm{q}$. The cost aggregation can efficiently be computed by traversing along each aggregation path, beginning at the image border, successively accumulating the costs and storing the cost for each pixel $\mathrm{p}$ in the aggregated cost volume $\mathcal{C}_\mathrm{aggr}$:  
\begin{equation}
\label{eq:sgm}
\begin{aligned}
\mathcal{C}_\mathrm{aggr}(\mathrm{p}, d_\mathrm{p}) &= \mathcal{C}(\mathrm{p}, d_\mathrm{p}) \\ &+ \sum_{\mathrm{q}\in \mathcal{N}_\mathrm{p}}{
	\begin{cases}
    	\mathcal{C}(\mathrm{q}, d_\mathrm{q}) + P_1 ,& \text{if } \left|d_\mathrm{p}-d_\mathrm{q}\right| = 1\\
	    \mathcal{C}(\mathrm{q}, d_\mathrm{q}) + P_2, & \text{otherwise}.
	\end{cases}}
\end{aligned}
\end{equation}

\begin{figure}[!h]%
	\centering
	\subfigure[]{\includegraphics[width=0.49\columnwidth]{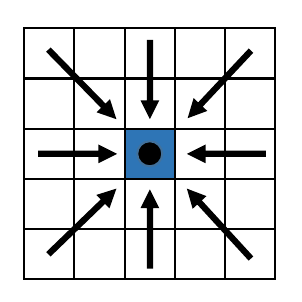}\label{fig:eight_paths}}\hfill%
	\subfigure[]{\includegraphics[width=0.49\columnwidth]{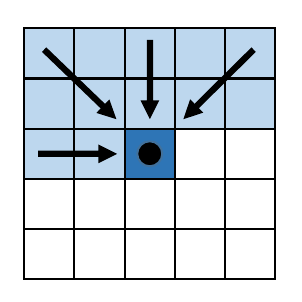}\label{fig:four_paths}}%
	\caption{Eight vs. four path aggregation. While the use of eight paths (a) require full image access for each pixel being aggregated, the use of only four paths (b) can efficiently be computed while streaming the pixel data and accumulating previously computed costs (light blue).}
	\label{fig:paths}
\end{figure}

As the data of the cost volume $\mathcal{C}$ is streamed through this processing stage, the amount of aggregation paths is reduced from eight to four paths as depicted in \Cref{fig:paths}. 

The \gls*{SGM} aggregation can efficiently be computed by simply buffering and accumulating the already computed costs. 
While the horizontal path only requires to buffer $d_\mathrm{max}$ values from the previous pixel, the vertical and diagonal paths require to buffer $3\times d_\mathrm{max}$ for each pixel of the previous row. 
The use of eight paths would require a second pass from back to front.
In order to avoid an overflow in the cost aggregation, the minimal path costs are subtracted after each processed pixel, as stated by \cite{Hirschmueller2008}.

From the aggregated cost volume $\mathcal{C}_\mathrm{aggr}$ the resulting disparity image $D$ can easily be extracted by finding the pixel-wise \gls*{WTA} solution $\hat{d}_\mathrm{p}$, \ie $D(\mathrm{p}) = \argmin_d \mathcal{C}_\mathrm{aggr}\left(\mathrm{p}, d\right)$

\begin{figure*}[!t]%
  \centering%
  \begin{overpic}[width=\textwidth]{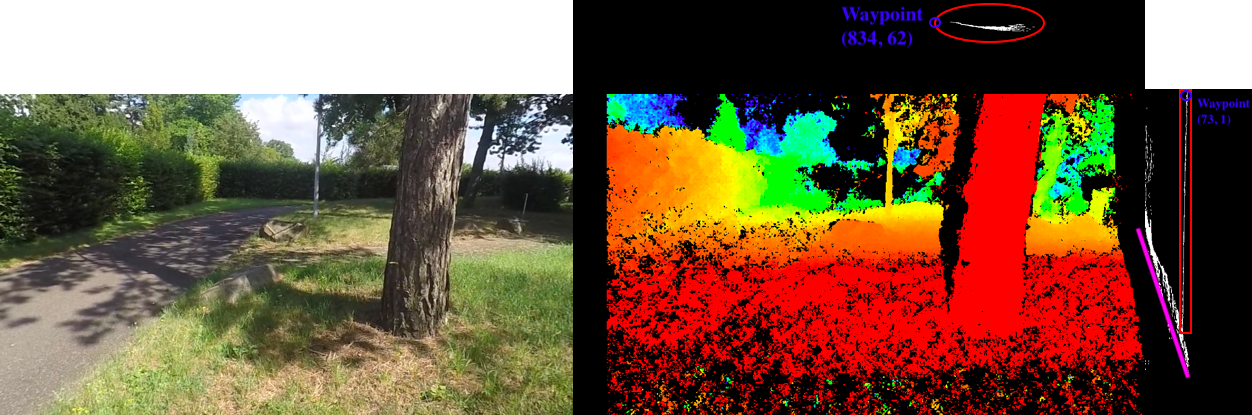}%
	\thicklines%
	\setlength{\fboxsep}{1pt}%
	\put(93.8,31.6){\fbox{\footnotesize U-Map}}%
	\put(93.8,32.0){\vector(-1,0){2.4}}%
	\put(93.8,29.0){\fbox{\footnotesize V-Map}}%
	\put(96.5,28,4){\vector(0,-1){2.4}}%
	\color{black}%
	\put(90.7,10){\line(0,1){10}}%
	\put(90.8,10){\line(0,1){10}}%
	\put(90.9,10){\line(0,1){10}}%
	\put(91.0,10){\line(0,1){10}}%
	\put(91.1,10){\line(0,1){10}}%
	\put(91.2,10){\line(0,1){10}}%
	\put(91.3,10){\line(0,1){10}}%
	\color{white}%
	\put(45.75,25.9){\line(1,0){45.7}}%
	\put(45.75,25.75){\line(1,0){45.7}}%
	\put(91.4,0){\line(0,1){26}}%
	\put(91.5,0){\line(0,1){26.1}}%
	\color{red}%
	\put(46.2,32.7){\vector(1,0){4}}%
		\put(49.5,31,4){$u$}%
	\put(46.2,32.79){\vector(0,-1){4}}%
		\put(47.0,28.7){$d$}%
	\put(91.92,25.6){\vector(1,0){2.4}}%
		\put(92.8,26.4){$d$}%
	\put(92.0,25.66){\vector(0,-1){4}}%
		\put(92.4,21.8){$v$}%
	\end{overpic}%
  \caption{Illustration of the U-/V-map. The image on the left shows the reference image from the left camera of the stereo system. On the right the coloured disparity image is depicted, together with the U-map on top and the V-map on the right. The depths in the disparity image are colour coded, going from red (near) to blue (far). The white contour of the U-map represents the tree and its width (red). The V-map encodes ground plane (highlighted in pink) and the height of the tree (red). Furthermore the U-/V-map depict the computed waypoint.}%
\label{fig:maps}%
\end{figure*}

\begin{figure*}[!ht]
  \centering
  \includegraphics[width=\linewidth]{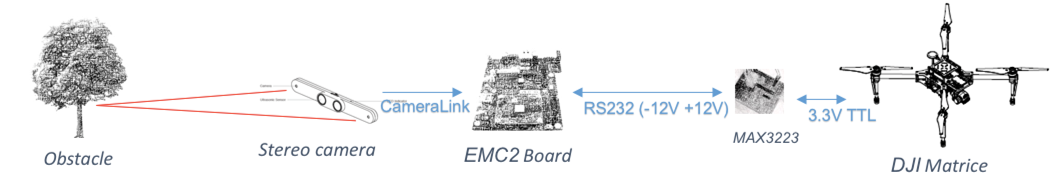}
  \caption{Schematic overview depicting interoperability of our approach with the DJI Matrice 100 \gls*{UAV} system.}
  \label{fig:uav_scheme}
\end{figure*}

\subsubsection{Left-Right consistency check}

Typical for stereo imagery occluded areas are not the same for both cameras due to different viewing perspectives onto the scene. Hence, it is not possible to compute disparities for areas which are occluded at least in one of the two cameras. 

In order to remove outliers, a common approach is to perform a left-right consistency check. As the name implies a disparity map $D_\mathrm{left}$ corresponding to the left image is compared to a disparity map $D_\mathrm{right}$ corresponding to the right image. For all disparities $d_\mathrm{left} \in D_\mathrm{left}$ the corresponding disparity in $D_\mathrm{right}$ is evaluated. If both disparities do not coincide, \ie $\left|D_\mathrm{left}(\mathrm{p}_x, \mathrm{p}_y) - D_\mathrm{right}(\mathrm{p}_x + d_\mathrm{left}, \mathrm{p}_y)\right| \leq 1$, the corresponding pixel in $D_\mathrm{left}$ is invalidated. This process typically requires the computation of an additional disparity map. However, $D_\mathrm{right}$ can be efficiently approximated by reusing the previously computed $\mathcal{C}_\mathrm{aggr}$ to compute the \gls*{WTA} disparities: $D_\mathrm{right}(\mathrm{p}) =\argmin_d \mathcal{C}_\mathrm{aggr}\left((\mathrm{p}_x + d, \mathrm{p}_y), d\right)$. 

\subsubsection{Median filter}

A final median filtering is employed in order to remover further outliers. Similar to the cost functions, the implementation of a $k \times k$ median filter on FPGAs requires $k$ line buffers.

\subsection{Obstacle Avoidance}

The second part of our system aims at the actual obstacle avoidance based on the previously computed data. The disparity map is used to estimate the depth of the depicted objects and detect obstacles in the flight path. Given that an obstacle is detected, our algorithm computes the shortest route around the object and manoeuvres the \gls*{UAV} around it. In the field of autonomous navigation there has been a lot of work for avoiding obstacles and navigating securely through an known or unknown environment. Nevertheless, the \TULIPP use-case aims at fast reaction times by keeping the weight and energy consumption as low as possible.

Our approach is therefore based on a reactive obstacle avoidance. The disparity information of the stereo images is used to build U-/V-Maps of the environment \citep{Labayrade2002, Li2014, Oleynikova2015}.
Hereby, the disparity image is transformed into a simpler representation in order to reduce complexity, to improve the robustness regarding inaccuracies in the measurements, and to simplify the obstacle detection. 
For every column of the disparity image a histogram of disparity occurrences is computed resulting in a map of size $W \times d_\mathrm{max}$ (cf. \Cref{fig:maps}).
This so called U-Map encodes the depth and width of each object in the disparity map and can be interpreted as a bird's eye view on the scene. 
Analogous to the U-Map, a V-Map of size $d_\mathrm{max} \times H$ is computed by creating a histogram of disparity occurrences for each row of the disparity image.
The V-Map reveals the ground plane, highlighted by the pink line in \Cref{fig:maps}, as well as the height of the objects at a given disparity. 

By applying a threshold filtering to the U-/V-Maps we transform them into binary maps, hereby suppressing uncertainties and revealing prominent objects. We further filter the results by applying dilatation and a Gaussian filtering to the maps. To be independent from the cluttered horizon only a region of interest is considered. In the next step we extract the largest contours together with their centroids from the U-/V-Maps analogous to the approach in \citep{Suzuki1985}. Then we perform an abstraction step by drawing ellipses in the binarized U-Map and rectangles in the binarized V-Map around the extracted contours as shown in \Cref{fig:maps}. This corresponds to simplified cylindrical object models of the obstacles in 3d space. If the obstacles have a critical distance to the \gls*{UAV}, or algorithm tries finds the shortest path to the left, right, up or down. The algorithm is executed periodically in order to exclude false positive detection. A wrong decision due to a false detection in one frame can be corrected in the next frame.

\begin{figure*}[!t]%
	\centering
	\subfigure[Refence image]{\includegraphics[width=0.32\textwidth]{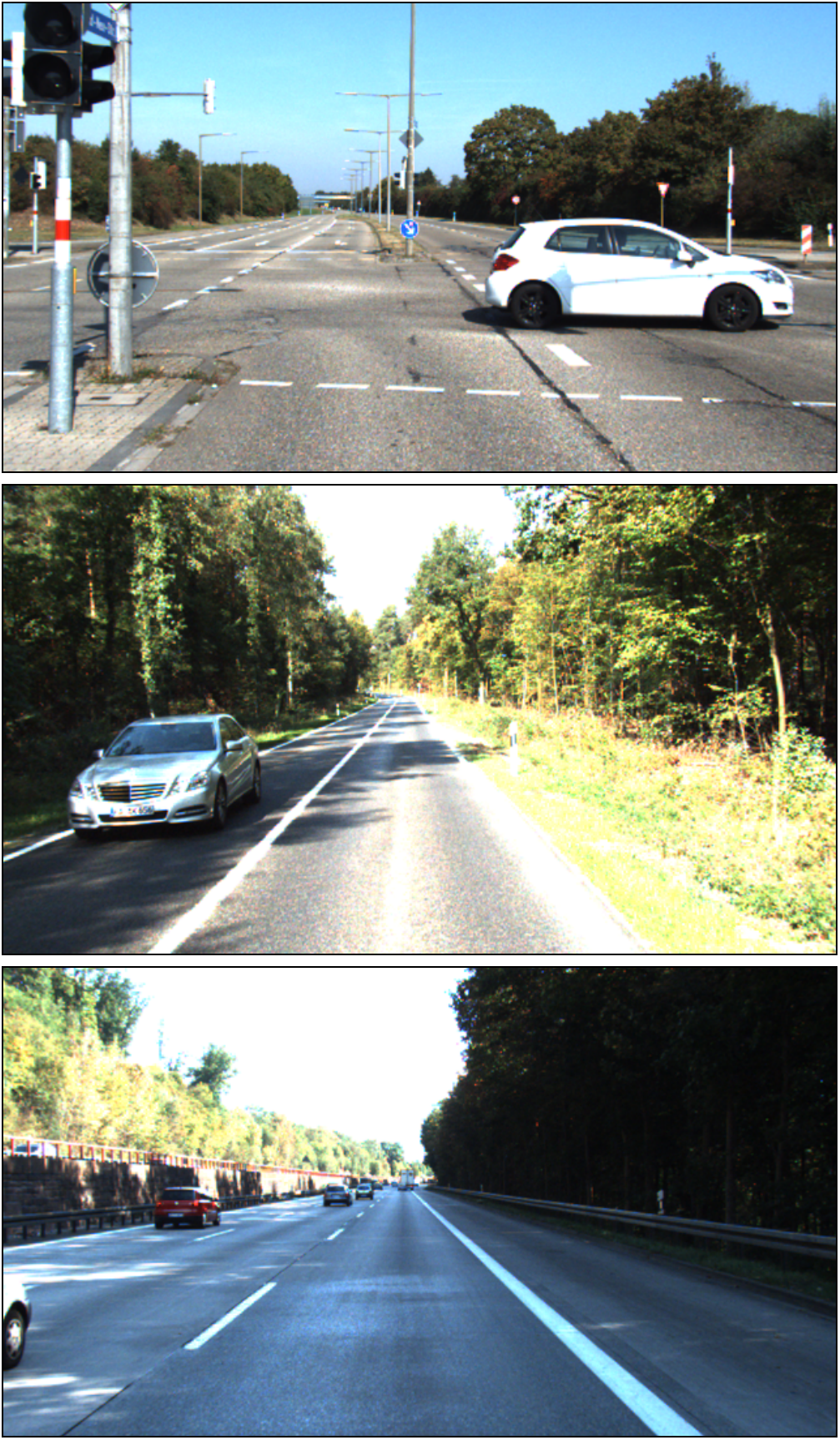}\label{fig:kitti_ref}}\hfill%
	\subfigure[Disparity image produced by using \gls*{SAD}]{\includegraphics[width=0.32\textwidth]{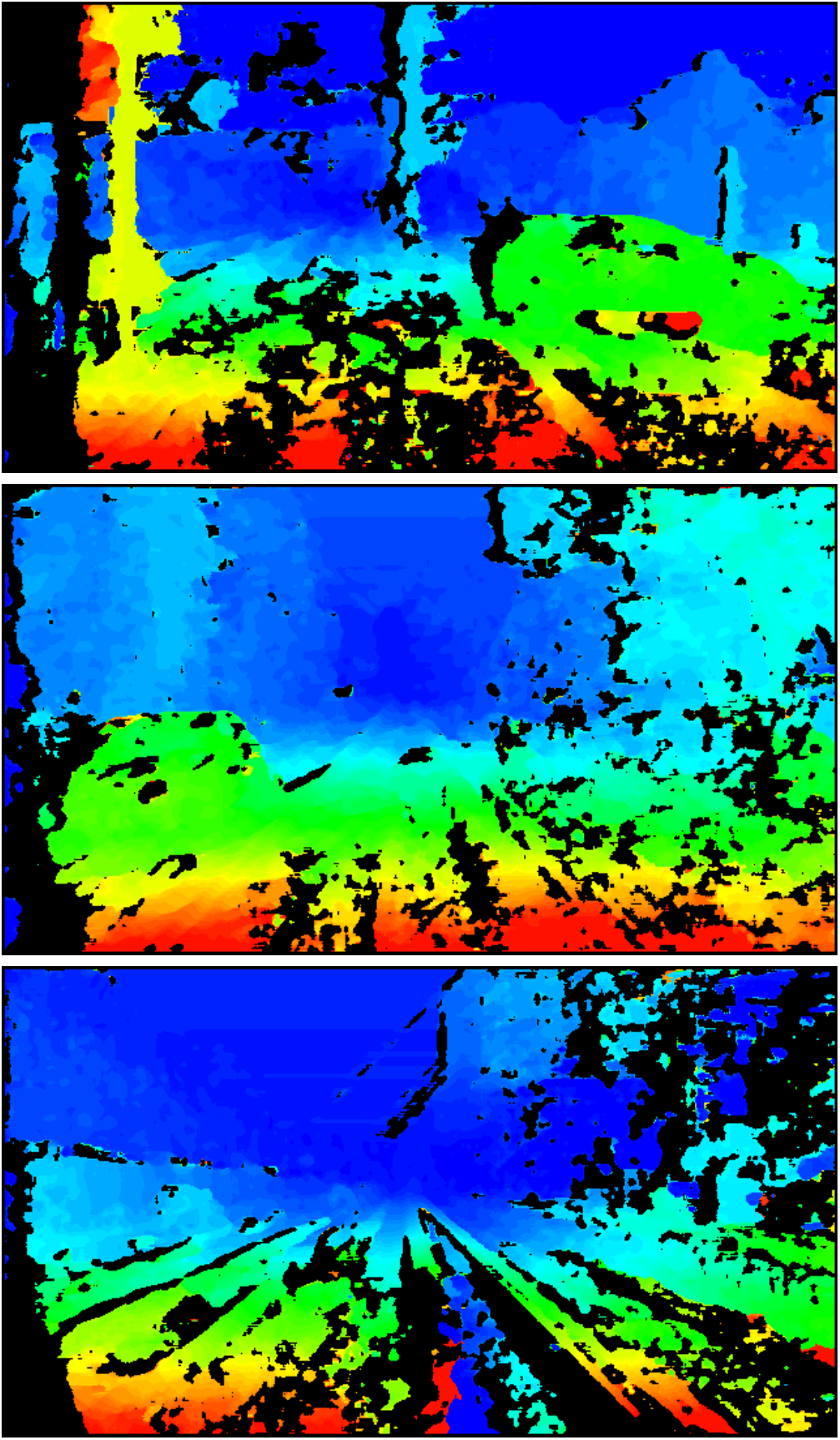}\label{fig:kitti_absdiff}}\hfill%
	\subfigure[Disparity image produced by using \gls*{CT}]{\includegraphics[width=0.32\textwidth]{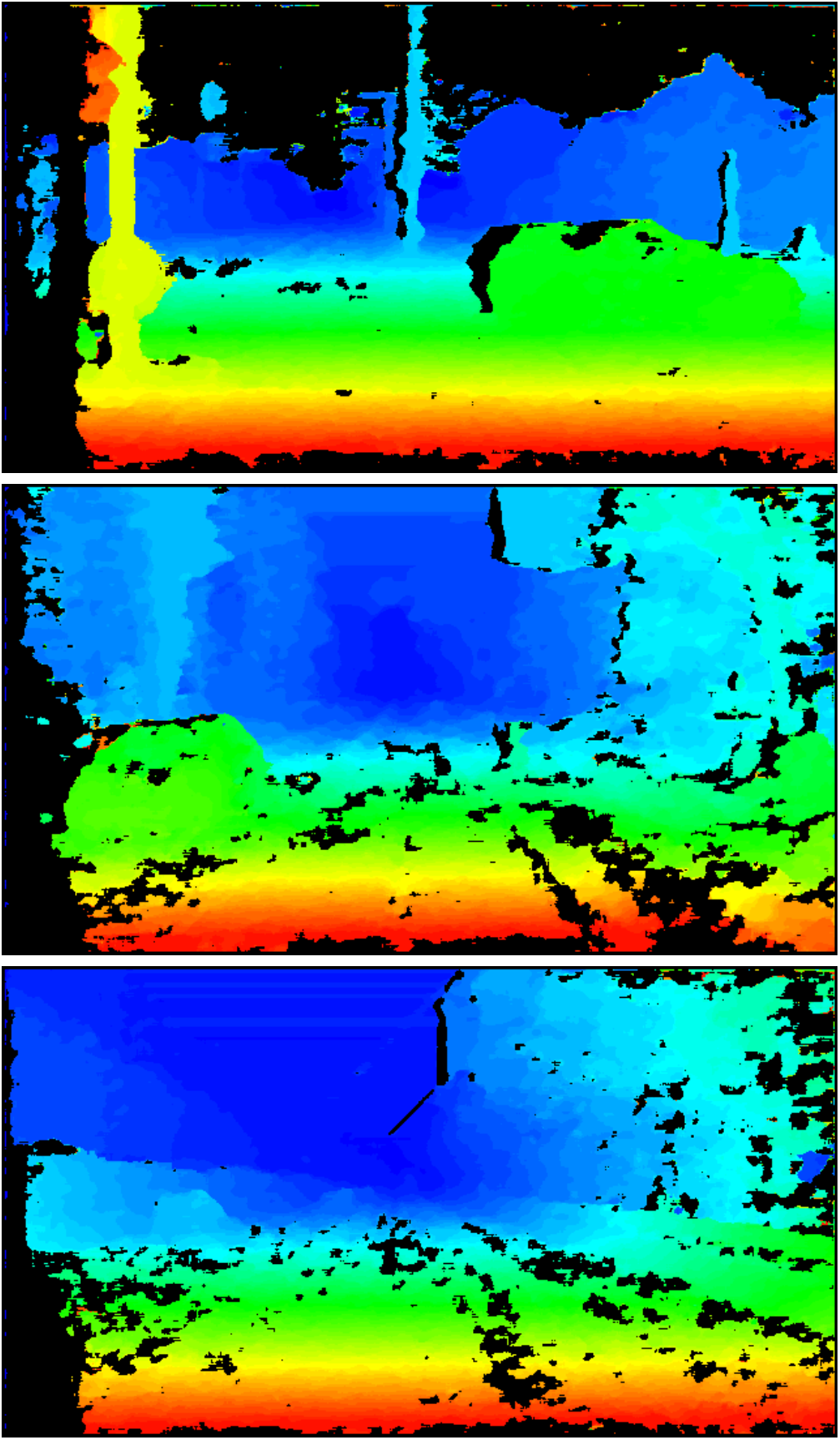}\label{fig:kitti_census}} %
	\caption{Qualitative results achieved by our algorithm for disparity estimation on the \KITTI Stereo 2015 Benchmark \citep{Menze2015CVPR}. 
	The disparity is colour coded, going from red (large disparity, near) to blue (small disparity, far). The black areas correspond to undefined disparity areas.}
	\label{fig:results_kitti}
\end{figure*}

\section{EXPERIMENTS}
\label{sec:eval}
\sloppy

We implemented and deployed our system for image-based obstacle avoidance on an embedded Xilinx Zynq Ultrascale+ with an ARM Cortex-A53 quad-core CPU and a FinFET+ FPGA. It is operated on the EMC$^2$-DP carrier board from Sundance. 
As input we use two industrial cameras from Sentech which are connected by CameraLink to the carrier board. 
Our system uses the \gls*{MAVLink} protocol to control the \gls*{UAV} and execute the evasion manoeuvre. 
We use the serial port to connect the carrier board to the flight controller of the \gls*{UAV} (\cf \Cref{fig:uav_scheme}). 

As already stated, the disparity estimation algorithm is run on the FPGA, while the obstacle avoidance is processed on quad-core CPU. 
\Cref{tab:runtime} shows runtime measurements of different parts of our system. 
A thorough description on the experiments done and the results achieved is given in the following two sections.

\begin{table}[!h] 
\caption{Runtime measurements of the proposed system achieved on a Xilinx Zynq Ultrascale+.}
	\label{tab:runtime}
	\centering
    \begin{tabular}{| l | r | r |}
    \hline
    Operation & \multicolumn{1}{|l|}{Time (s)} & \multicolumn{1}{|l|}{Average time (s)} \\ \hline\hline
    Disparity map & 0.0314 - 0.0361 &  0.0345 \\ 
    Obstacle avoidance & 0.0027 - 0.0110 & 0.0042 \\ \hhline{|=|=|=|}
    $\Sigma$ & 0.0341 - 0.0471 &  0.0387 \\ \hline
    \end{tabular}
\end{table}

\subsection{Real-time disparity estimation}
We have used Xilinx Vivado \gls*{HLS} to synthesize our algorithm from C/C++ into VHDL code and deployed on the FPGA running at 200\MHz. 
The FPGA of the Ultrascale+ provides 154\kilo logic cells. 
\Cref{tab:resource_util} holds figures on how many resources our algorithm utilizes on the Ultrascale+. 
We have configured our algorithm to use a 5$\times$5 support region for the \gls*{SAD} and \gls*{CT} similarity measure, as well as for the final median filtering. 
The disparity sampling range is set to $d = [0,60)$. 
The \gls*{SGM} penalties were set to $P_1 = 200$ and $P_2 = 800$ for the \acrlong*{SAD} cost function, and to $P_1 = 8$ and $P_2 = 32$ when using the Hamming distance of the \acrlong*{CT}.  
Our full pipeline achieves an average processing speed of 29\Hz at a frame size of 640$\times$360 pixels. 
The latency of our synthesized code is calculated to be 28.5\ms. 
Experiments have shown that a reconfiguration of the FPGA to run at only 100\MHz leads, to a processing speed of \approxi 20\Hz and a latency of 53.2\ms.

\begin{table}[!ht] 
\caption{Resource utilization with respect to the Zynq Ultrascale+ of the \gls*{HLS} code for disparity estimation.}
	\label{tab:resource_util}
	\centering
    \begin{tabular}{| l | r | r | r |}
    \hline
    Resource 	& 		\multicolumn{1}{|l|}{Used} 	& 	\multicolumn{1}{|l|}{Available}	& \multicolumn{1}{|l|}{Utilization} \\ \hline\hline
	DSP48E 		& 		22 		& 	360 		&  6 \% \\ 
    BRAM\_18K 	& 		132 	& 	432 		& 30 \% \\  
    FF 			& 		12561 	& 	141120 		& 8 \% \\
    LUT 		& 		27063 	& 	70560 		& 38 \% \\ \hline
    \end{tabular}
\end{table}

To quantitatively evaluate the results achieved, we have adapted our processing pipeline to read pre-recorded and rectified imagery from memory. 
We have used the \KITTI Stereo 2015 benchmark \citep{Menze2015CVPR} for evaluation, cropping a \gls*{RoI} of size 640$\times$360 in order to not alter the algorithm.
\Cref{fig:results_kitti} shows exemplary results achieved by our algorithm.
While the middle column (\Cref{fig:kitti_absdiff}) contains the results achieved when using the \gls*{SAD} similarity measure, the results in the right column (\Cref{fig:kitti_census}) reveal that a use of the \gls*{CT} allows a more robust matching between the input images and thus yielding results that are more accurate and hold less pixels that are invalidated due to inconsistencies in the left-right check. 
These observations are justified by the contents of \Cref{tab:result_kitti}, holding the quantitative measurements \wrt to the benchmark. The performance of our algorithm on subsequence of the benchmark is given at: \url{https://youtu.be/gzIFqUmqM7g}.

\begin{table}[!ht] 
\caption{Quantitative results achieved by our algorithm for disparity estimation \wrt to the \KITTI Stereo 2015 benchmark, depending on the similarity measure used. A pixel is considered to be estimated correctly if the disparity $<$\,3\px \wrt to the ground truth.
The density specifies percentage of pixels in the resulting disparity image for which the disparity was computed.}
	\label{tab:result_kitti}
	\centering
    \begin{tabular}{| l | r | r |}
    \hline
			Cost function & Density & Correct pixels \\
			\hline
			\hline 
			\gls*{SAD} & 64.1 \% & 76.0 \% \\ \hline
			\gls*{CT}  & 74.2 \% & 95.4 \% \\ \hline
    \end{tabular}
\end{table}

\subsection{Obstacle Avoidance}
\begin{figure}[!ht]%
\centering
	\subfigure[Screenshot of the DJI flight simulator software]{\includegraphics[width=0.49\columnwidth,frame]{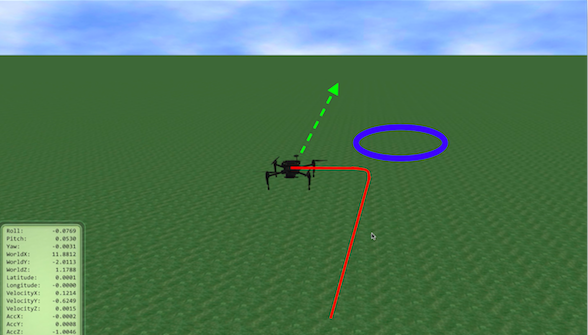}\label{fig:sim_dji}} \hfill%
    \subfigure[Screenshot of the Microsoft AirSim simulator software]{\includegraphics[width=0.49\columnwidth,frame]{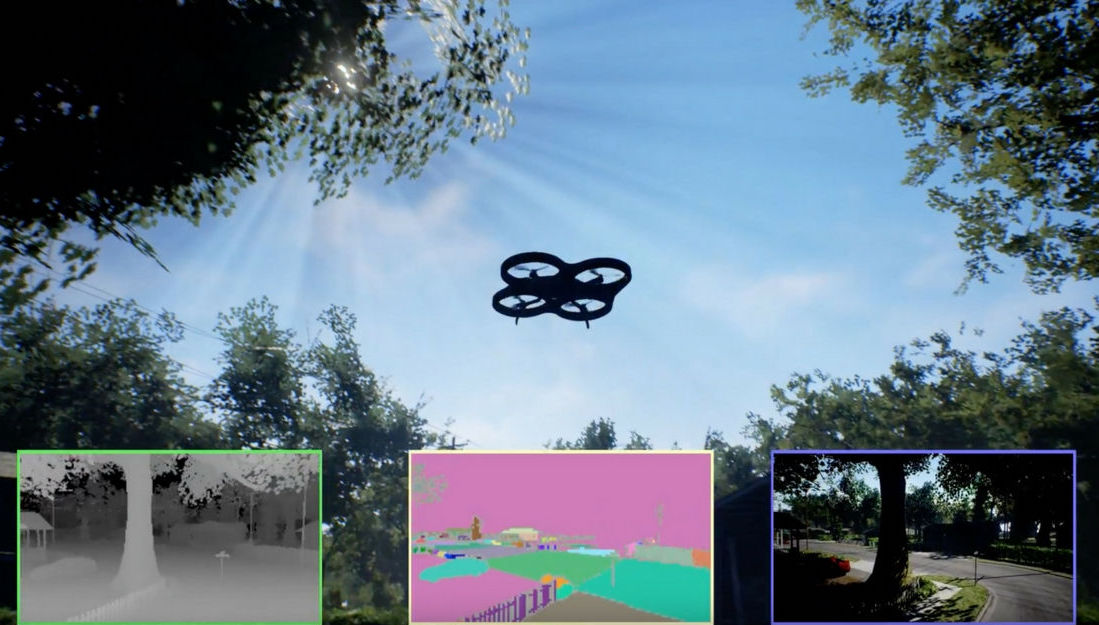}\label{fig:sim_airsim}} %
    \caption{The two flight simulators used for Hardware-in-the-Loop testing. While the main view in (b) shows a third-person view point on the scene, the three smaller displays at the bottom reveal the disparity map, an object segmentation and camera image from the \gls*{UAV} point of view.}%
\label{fig:sim}
\end{figure}

We integrated the obstacle avoidance system into two different \gls*{UAV} systems, namely a Pixhawk based system and the the DJI Matrice 100 as shown in \Cref{fig:uav_scheme}. 
To validate the integrity of our algorithm we used Hardware-in-the-Loop testing in conjunction with a \gls*{UAV} flight simulators from DJI and Microsoft AirSim \citep{Shah2018airsim}, as shown in \Cref{fig:sim}.
Both flight controller were connected through their USB interface to the host PC running the flight simulator. 
With the Hardware-in-the-Loop connection the application for the \gls*{UAV} is the same as in real environment, but instead of controlling the motors the flight of the \gls*{UAV} is simulated.

For the DJI based system we used the proprietary flight simulator, which is very minimalistic but provides all functionality we need. 
The tests were executed in autonomous flight mode. Hereby the \gls*{UAV} takes off to a height of one meter and accelerates in forward direction. 
If there is an object in the flight path, the UAV finds the shortest path to the edge of the object and accelerates to the left, right, up or down. 
If the object has moved out of the field of view the \gls*{UAV} starts accelerating in forward direction again. This setup allowed us to test the reaction of the our algorithm \wrt to incoming obstacles. 
We provide a visual demonstration of the simulation at: \url{https://youtu.be/pMDMTUCVwKc}

For further improvement of the algorithm for obstacle avoidance it is necessary to use appropriate simulation software. 
Therefore we changed to the Pixhawk flight controller which is supported by the Microsoft AirSim simulator. 
The simulator is based on the Unreal Engine which provides a photo-realistic environment. 
It allows to generate depth maps of the environment which allowed us to test the performance of our obstacle avoidance algorithm directly. 
However, these depth maps are synthetically generated and thus do not resemble realistic results achieved by our disparity estimation in a real world scenario. 
The AirSim simulator allows to thoroughly test and analyse the behaviour of our obstacle avoidance in different scenarios.

\section{CONCLUSION \& FUTURE WORK}
\label{sec:conclusion}

\sloppy

In this paper we present a system for real-time on-board obstacle avoidance for \gls*{UAVs} based on an embedded stereo vision approach. 
It uses imagery from a stereo camera system for a real-time disparity map estimation of the environment in front of the \gls*{UAV},
which is then used for the adaptation of the flight path.

The algorithm for image-based disparity estimation, which adopts the \acrlong*{SGM} approach, is optimized for the deployment on an embedded FPGA. 
Hereby the strategy for optimizing the \gls*{SGM} energy function is adapted in order to account for the strengths of FPGAs to process data streams.
This is done, among other, by using four aggregation paths rather than the commonly used eight.
It runs on the FPGA, which is operating at 200\MHz and achieves a processing speed of 29\Hz and a latency of 28.5\ms at a frame size of 640$\times$360.
This is significantly less powerful than state-of-the-art systems, however, since this work is part for the \TULIPP project, which is aiming, amongst other things, at reducing the gap between application development and hardware optimization, we focus on a user friendly development by implementing the algorithm in C/C++ and porting it with the use of \acrlong*{HLS} to the FPGA.

The second part of our system transforms the previously computed disparity map 
into simpler representations, the so called U-/V-Maps,
which are used for obstacle detection.
Having found an obstacle in the flight path, the algorithm computes the shortest path and manoeuvres the \gls*{UAV} around it.
Due to the low computational complexity of our reactive obstacle avoidance algorithm, it can be deployed on the embedded CPU without any major performance loss.

We deployed our system on an embedded Xilinx Zynq Ultrascale+ with an ARM Cortex-A53 quad-core CPU and a FinFET+ FPGA.

When computing disparity maps we compared two commonly used similarity measures for image matching, namely the \acrlong*{SAD} and the Hamming distance of the \acrlong*{CT} and found out that \acrshort*{CT} outperforms the \acrshort*{SAD}.
The accuracy of the algorithm for disparity estimation was evaluated on the \KITTI Stereo 2015 benchmark achieving $95.4 \%$ of correctly estimated pixels at a density of $74.2 \%$ \wrt to the groundtruth when using the \acrshort*{CT} as the similarity measure.
The integrity of the reactive obstacle avoidance algorithm is evaluated by using Hardware-in-the-Loop testing in conjunction with two flight simulators.

Our current work is focused on carrying out experiments in the real-world environment with the EMC$^2$-DP carrier board, together with the Ultrascale+ and the camera system mounted to the \gls*{UAV}.
In parallel, 
we plan to evaluate our stereo algorithm \wrt energy consumption on different embedded hardware architectures, namely FPGA, CPU and GPU.

\section*{ACKNOWLEDGEMENTS}
\label{sec:acknowledgments}

The \TULIPP project is funded by European Commission under the H2020 Framework Programme for Research and Innovation under grant agreement No 688403. 

{
	\begin{spacing}{0.9}
		\bibliography{tulipp-isprs-tc1_biblio} 
	\end{spacing}
}

\end{document}